\documentclass[12pt]{article}

\usepackage[utf8]{inputenc}
\usepackage[english]{babel}
\usepackage[T1]{fontenc}
\usepackage{multirow}
\usepackage{graphicx}
\usepackage[linesnumbered,ruled,vlined]{algorithm2e}
\usepackage[a4paper, margin=2.7cm]{geometry}
\usepackage{graphicx}
\usepackage{subcaption}
\usepackage{enumerate}
\usepackage{amsmath, amsthm, amsfonts, amssymb}
\usepackage{mathrsfs}           
\newcommand\tab[1][1cm]{\hspace*{#1}}

\usepackage[HTML]{xcolor}
\usepackage[export]{adjustbox}
\definecolor{MyColor}{HTML}{000000}

\usepackage{float}

\usepackage[colorlinks=true,linkcolor=blue,citecolor=blue,urlcolor=blue,breaklinks]{hyperref}

\SetKwInput{KwInput}{Input}                
\SetKwInput{KwOutput}{Output}              

\usepackage{bbm} 

\date{}

\begin{document}

\begin{flushleft}

\begin{center}

{\Large
\textbf{TPM: Transition Probability Matrix - Graph Structural Feature based Embedding} \par
}

\bigskip

Sarmad N. MOHAMMED\textsuperscript{1,2,*},
Semra G\"UND\"U\c{C}\textsuperscript{2}
\\
\bigskip

{ \footnotesize
$^{1}$  Computer Science Department, College of Computer Science and Information Technology, University of Kirkuk, Kirkuk, Iraq
\\
$^{2}$ Computer Engineering Department, Engineering Faculty, Ankara University, Ankara, Turkey

\bigskip

* sarmad\_mohammed@uokirkuk.edu.iq
}
\end{center}

\end{flushleft}

\begin{abstract}
In this work, Transition Probability Matrix (TPM) is proposed as a new method for extracting the features of nodes in the graph. The proposed method uses random walks to capture the connectivity structure of a node's close neighborhood. The information obtained from random walks is converted to anonymous walks to extract the topological features of nodes. In the embedding process of nodes, anonymous walks are used since they capture the topological similarities of connectivities better than random walks. Therefore the obtained embedding vectors have richer information about the underlying connectivity structure. The method is applied to node classification and link prediction tasks. The performance of the proposed algorithm is superior to the state-of-the-art algorithms in the recent literature. Moreover, the extracted information about the connectivity structure of similar networks is used to link prediction and node classification tasks for a completely new graph.
  
  
\end{abstract}

\section{Introduction}
\label{intro}
\tab The network concept is a mathematical model of structures where elements, the nodes, are connected through the edges. Network connections are because of semantic or geometric relations among the nodes. Entities of complex networks, such as individuals, molecules, neurons, and computers, are represented by nodes, while their relations or interactions make up the edges. These structures, the networks, are observed in diverse areas, and they are primary mathematical tools for modeling complex systems~\cite{albert2002statistical, barabasi2013network}. Common properties of complex systems, such as clustering~\cite{han2017linking}, forming communities~\cite{cherifi2019community}, and new links~\cite{chi2019link} are all in the interest of researchers of different disciplines, as the common network structures are observed in diverse areas, such as physical, biological, and social sciences. For this reason, networks have opened new opportunities for a better understanding of the underlying dynamics of diverse problems of ever-growing complexity. Moreover, new challenges appear as the sizes of the networks and the complexity of the problems of interest increase. The challenges are manifold. Among various challenges in data mining, getting the correct picture of the relations in large networks profoundly exhibits itself. The problems related to network mining, such as community detection~\cite{fortunato2010community, cetin2022new}, node classification~\cite{bhagat2011node}, link prediction~\cite{martinez2016survey}, structural network analysis~\cite{kossinets2006empirical}, and network visualization~\cite{tamassia2013handbook}, are being studied.

\tab Nodes in complex networks carry highly structured information. Considerable effort has been devoted to processing this information over the last few years. The issue is to correctly identify the feature vector representation of the nodes and edges. Old-fashioned traditional solutions offer hand-engineering through expert knowledge. Because this is for special cases, it is challenging to generate different tasks. For this purpose, various graph embedding approaches ~\cite{sun2020network} have been introduced. Graph embedding is a map between high-dimensional, highly structured data and low-dimensional vector space that preserves the structural information of all nodes in the network. The embedding process gains importance due to the growing number of applications that benefit from network data in a broad range of machine learning domains, such as natural language processing (NLP), bioinformatics~\cite{zhang2021graph}, social network analysis~\cite{mele2021structural,vencalek2021depth}, and recently as building blocks of reinforcement learning algorithms~\cite{battaglia2018relational, zhang2020deep}.

\tab In-network-related problems, node classification, finding relations, and predicting non-apparent connections among the entities are the crucial steps for approaching the problem. Hence, the extraction of implicit information present in the network and identifying the existing links is the base requirement of all network-related problems. The link prediction comprises inferring the existence of connections between network entities based on the properties of the nodes and observed links~\cite{liben2007link}. The purpose is to label an unlabeled node by evaluating a labeled node. In a link prediction task, the purpose is to predict a possible link between two nodes by considering the existing links in the network. The techniques designed to solve node classification and link prediction problems are also used in modeling new networks and predicting network evolution mechanisms~\cite{pavlopoulou2017predicting, khafaei2019tracing}.

\tab It is common practice in social networks and recommendation systems that two entities with similar interests are more likely to interact. Hence, commonly shared empirical evidence shows that similar nodes are likely to interact. Palla et al.~\cite{palla2005uncovering} observed that nodes tend to form connected communities. This observation has led to an accepted similarity definition as the amount of relevant direct or indirect paths between nodes. Therefore the challenge in node identification and link prediction is to define similarities in network entities. The nodes carry geometric identities, which are the reflections of network topology. Besides these topological characteristics, some networks also provide homophily identities for the nodes. This information has applicability among a wide range of different networks. A better understanding of the network domain and the existence of semantic information together with the geometric connectivity helps define the node similarity and increases the learning algorithm's efficiency.

\tab In node classification and link prediction, two main obstacles are the lack of information and the size of the real-world networks. Real-world networks consist of millions of nodes, and the number of links is a large multiple of the number of nodes. For such large structures, only geometric information may not be sufficient for node identification. This is particularly true in very sparse networks. The remedy, found in some algorithms, is to extend the network region surrounding the node. More sites are included in the information-gathering process by increasing the region to collect information. Here another bottleneck comes into play. Such algorithms are computationally expensive and can only apply to small networks. The amount of information collected from the network, the algorithm's complexity, and the techniques to collect information must be balanced.

\tab This work introduces a new scalable unsupervised node embedding algorithm to calculate the transition probability between the neighboring sites. The transition probability carries the characteristic connectivity structure of the region around the node in concern. The idea is to capture the local connection structure of a node by visiting a few close neighbors. This approach saves computation time, increases performance, and enables information collection without going deep into a graph. This is achieved by starting a random walk from each node in a predefined number of steps. Then random walks are converted into anonymous walks to extract the variety and richness of connections around the nodes. This approach focuses on capturing the neighboring structural patterns rather than the node's identity. These anonymous walks preserve the local structural information and are used in calculating the transition probability matrix (TPM) elements. The transition probability matrix defines the probability of being in neighboring nodes at each time step. The transition matrix elements include the probability of reaching the predecessor, new, or already visited node at each time step. This information creates a fingerprint of how dense or sparse local ties are.

\tab In the proposed work, the elements of the transition probability matrix are used as feature vectors for each node. The TPM method makes two main contributions to the literature. The first is to produce a feature vector representation using an anonymous walk and offer a new embedding method. The second one is to use this extracted information from sample networks in a new network with a similar topological connectivity structure. The idea is to use the similarity of connections without considering the node's details. In traditional classification problems, data is used to train the algorithm, and predictions are made on the same network, of which information is already used as a part of training data. To the best of our knowledge, the proposed algorithm is the first in the literature to test the performance of the embedding process on a completely different graph except for having the same topological connectivity structure. The power of the proposed algorithm comes from the generality of the local connectivities, demonstrating the similarity of the topological structures of the networks. Hence, structural equivalence (roles of nodes in the network) dominates the unavailable features to find the similarities needed for classification and prediction tasks. Nodes' inner structures or identifications are hidden in many real-world cases. So it is essential to use topological features in the embedding process. Despite being unique to a node in the given network, the topology of the real-world networks allows the node feature vector to be used in different similar networks for prediction. In this sense, the proposed algorithm exhibits possibilities for various applications. The possibility of obtaining information from a similar network for predictions on the new networks is the most general characteristic that makes the proposed algorithm superior to random walks or graph convolutional network-based models for many real datasets.

\tab The rest of the paper is organized as follows: the next section briefly describes the proposed embedding algorithm TPM. In Section 3, we experimentally test TPM on various network analysis tasks over three citation networks and examine the parameter sensitivity of our proposed method. Finally, our conclusions and future works are presented in section 4.

\section{Model}
\label{Method}

\tab Networks are mathematically represented by graphs $G(V, E)$. Graphs consist of a set of nodes ($V$) and edges ($E$) connecting nodes. Assuming no multiple connections exist among the nodes, a network of $N$ nodes can have at most $E_{Max} = N(N-1)/2$ undirected edges. Here, $E_{Max}$ is the maximum number of possible connections. The network structure with $E_{Max}$ connections is called "fully connected". Apart from the fully connected networks, all network topologies possess a number of connections less than $E_{Max}$. For any specific network, its topology limits the number of edges. Apart from the fully connected networks, each network has a characteristic pattern of connections around the vertices, making classification of the network topology possible. Depending on degree distribution, there are mainly different connectivity structure power law (scale-free) and Erdos-Renyi (Random) networks.

\tab The set of known edges $E$ is considered positive (existent), while the set $E_{Max} - E$ edges constitute the set of non-existent edges and are called negative edges. The unique local connectivity structure of the given network plays the most crucial role in node characterization. Node features are representations of relations among the sites in a given neighborhood radius. At this point, there exist different recipes and methods of node feature extraction (\cite{xu2021understanding} for a survey). Different feature extraction algorithms determine the features of a given node with relative success, computational complexity, and computational expense. The success in determining the features of the nodes is essential since it is used as an input for more complicated constructions, such as the process of link prediction and community detection. The link prediction is an identification process of the possible candidates of positive edge sets in the set of non-existing edges, $E_{Max}-E$. The prediction process proceeds by assigning weights, $w_{i,j}$ to all possible edges between the nodes, labeled $n_i$ and $n_j$. Weights are model-defined similarity relations between the neighboring nodes. The relations are determined by the rules based on the features of the nodes.  The possible positive edges are predicted according to their weight values; the higher the score, the more likely an edge exists.

\tab The node feature extraction algorithms commonly employ random walks. Notably, highly cited DeepWalk~\cite{perozzi2014deepwalk} and node2vec~\cite{grover2016node2vec} methods base their feature extraction algorithms on creating large sequences of random walks, in a similar setting to sentences in natural language processing. Words and sentences correspond to nodes and random walks, respectively. The relational information between the node in concern and neighboring nodes is obtained from random walk sequences. Apart from these two well-established algorithms, a new, random walk-based approach for learning entire network representation has been introduced~\cite{ivanov2018anonymous} recently. In this new approach, random walks are converted to anonymous walk sequences by relabeling nodes by their occurrences in the walk.

\tab In the present work, anonymous walks are a good candidate for extracting topological features of the whole network and are used to extract feature vectors of the nodes. The proposed method showed that learning the node representation is possible by extracting local connectivity information as node features via anonymous walks.

\subsection{Anonymous Random Walks}
\tab An anonymous random walk is a process of relabelling the nodes. In a local random walk sequence, nodes are labeled according to their occurrence to obtain an anonymous walk sequence. A random walk of length $m$ starting from the node $v_i$ is a set of node labels, $ R_w = (v_1, v_2, \dots, v_m ) $ where all $v_i\; \in\; V $. The anonymization process is a mapping between the actual node labels and their occurrences in the sequence, $A_w= (f(v_1) \dots ,f(v_m))$. This anonymization process provides encapsulated information to reconstruct the local structure of the graph without requiring global node labels. Micali \& Zhu~\cite{micali2016reconstructing} have shown that by using anonymous random walks of length, $m$, starting from a node $v_i$ are used to reconstruct the immediate neighborhood of the node $v_i$. Therefore, using anonymous random walks, local topological features of nodes provide consistent information to construct a global structure of the graph.

\subsection{Problem Formulation}
\tab The behavior of the random walks represents the local connectivity structure of any node $v_i$. The random walk sequences contain new nodes at almost every step if the region is loosely connected. The probability of returning to the already visited sites increases if the neighborhood is densely connected. This information gives a unique picture of the connectivity structure near node $v_i$. Therefore, the transition probability is a good candidate for the extraction of the topological features of the nodes. The proposed algorithm aims to extract a map of the local topological structure of a given node using the transition probability. The proposed node embedding is based on calculating transition probability from any node within a certain number of steps around the node $v_i$. The probability of reaching a neighboring node is proportional to the degree of the node in concern. Hence, from a set of anonymous random walks, all starting from node $v_i$, the obtained distribution of reaching a new node or moving back to one of the already visited nodes at a given time step constitutes the elements of the transition matrix. The feature embedding vector of a node $v_i$ is the transition probability matrix obtained from the transition matrix.

\subsection{Obtaining Transition Matrix-Conceptual Considerations}
\tab Using random walks makes it possible to obtain a sequence of nodes in the graph. The positions of the nodes in this sequence will be related to the connectivity structure of the graph and keep rich information about the local neighborhood. The more distance we travel through the graph, the more global information we obtain. Local connectivity structure and variety of neighbors of any node are significant  for some tasks, graph recovery, reconstructing topology~\cite{micali2016reconstructing,ccetin2022new}, etc. For this reason, a limited length of random walks $m$ starts from each node is done and converted to $m$ step anonymous walks. The initial node, $v_i$, is considered the zeroth node, $a_0$, in the anonymous walk. The next node in the sequence is one of the nearest neighbors of node $v_i$, reached in the first step of the random walk. Each neighbor of $v_i$ has the same probability of being visited, which is related to the degree ($d_i$) of $v_i$. The same process continues for the next steps of random walks, so the underlying topological structure of the neighborhood of the visited nodes is embedded into the details of the next step.

\tab In principle, depending on the connectivity structure of one of the previously visited nodes, a new unvisited node or previous node can be reached at the next step. The only condition to reach any previously visited node in the next steps is to have a direct connection to that node. Hence, the high probability of reaching an already visited node is an indication of the densely connected local region.  

\tab Let a random walk $R_w = (v_1, v_2, \dots, v_m )$ where all $v_i\; \in\; V $. The corresponding anonymous walk is obtained by calculating the position of each node in the walk. So if an already visited node is visited at any step of the walk, the position index value that appears for the first time is used \cite{ivanov2018anonymous}.

\begin{figure}[h]
   \centering
  \includegraphics[scale=1.3]{"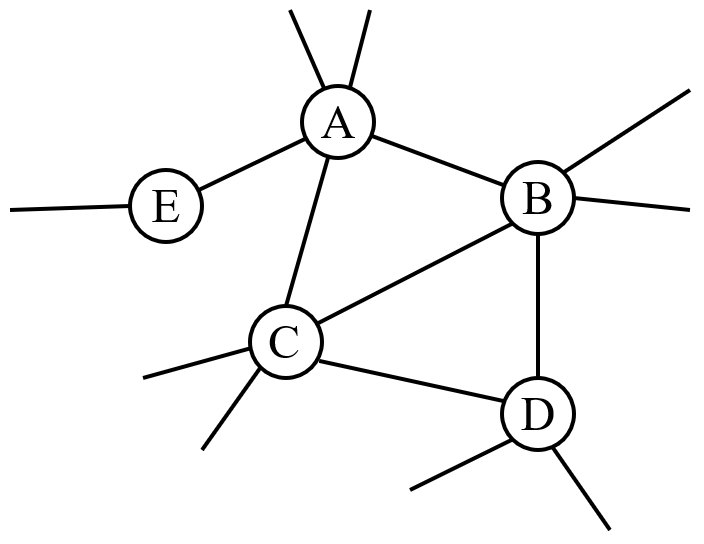"}
\caption{Sample subgraph.}
   \label{SampleGraph_1}
  \end{figure}
  
\tab In this sample graph (Figure~\ref{SampleGraph_1}), let's start a set of random walks with length $4$ from node $A$ and convert them to anonymous walks.

\begin{description}
\centering
\item 
$
R_{w1} : A \rightarrow B \rightarrow C \rightarrow A \rightarrow E \hspace{5em} 
A_{w1} : a_0 \rightarrow a_1 \rightarrow a_2 \rightarrow a_0 \rightarrow a_3
$
\item 
$
R_{w2} : A \rightarrow C \rightarrow D \rightarrow C \rightarrow A \hspace{5em} 
A_{w2} : a_0 \rightarrow a_1 \rightarrow a_2 \rightarrow a_1 \rightarrow a_0
$
\item 
$ 
R_{w3} : A \rightarrow B \rightarrow D \rightarrow B \rightarrow A  \hspace{5em} 
A_{w3} : a_0 \rightarrow a_1 \rightarrow a_2 \rightarrow a_1 \rightarrow a_0
$
\end{description}

\tab The corresponding anonymous walks have information about the variety of neighbors and topological structures. For this reason, it is possible that two different random walks ($R_{w2}$ and $R_{w3}$) can produce the same anonymous walks. In a random walk for the next step to move, there are mainly three different possibilities:

\begin{enumerate}[(i)]  
\item $P_{new}$ \hspace{1em} visit a new node.
\item $P_{prev}$ \hspace{1em} visit (go back) to previous node.
\item $P_{already}$ \hspace{1em} visit an already visited node.
\end{enumerate}

\begin{figure}[h]
   \centering
  \includegraphics[scale=1.3]{"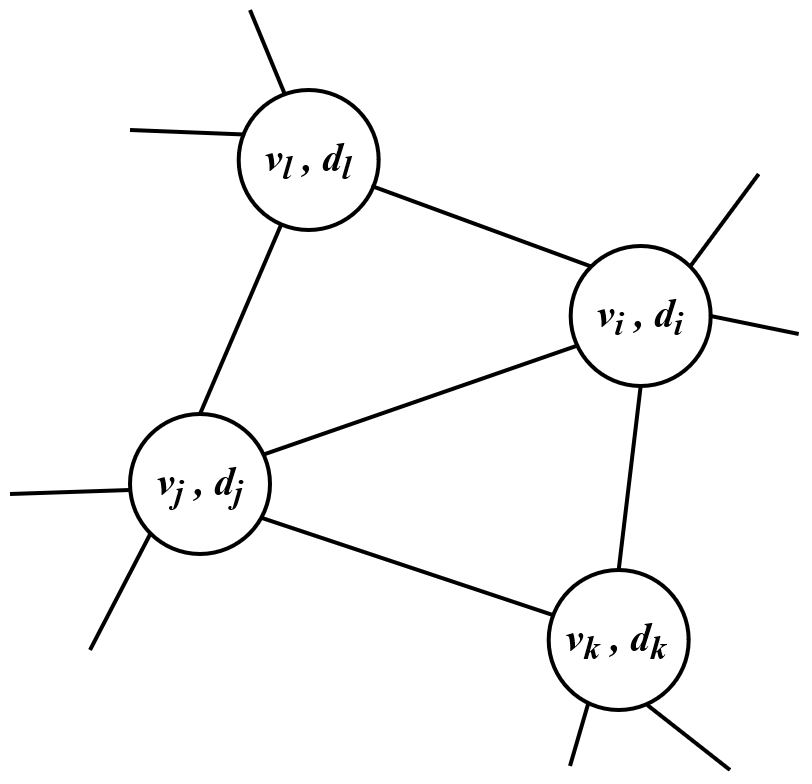"}
\caption{Sample subgraph, $d_i$ is degree of node $V_i$. }
   \label{SampleGraph_2}
  \end{figure}
  
\tab For a random walk starts from node $v_l$ in Figure~\ref{SampleGraph_2}, let's assume that first step is node $v_i$ and the second step is node $v_j$ ($v_l \rightarrow v_i \rightarrow v_j$). For the next step, possible choices are:

\begin{description}
\item $P_{j,k}$ \hspace{1em} to a new node
\item $P_{j,i}$ \hspace{1em} to a previous node
\item $P_{j,l}$ \hspace{1em} to an already visited node
\end{description}

\tab Let $P^{t}_{i,j}$ be the transition probability from node $i$ to node $j$ at time step $t$. Considering a three-step random walk. The possible sequence of the anonymous walk is,

\begin{table}[H]
\begin{tabular}{ccccc}
At step 0 \tab & with \tab & $P=1$ \tab & be in node $a_0$ \tab & $a_0$ \\ [1.2ex]
At step 1 \tab & with \tab & $P^{(1)}_{new}$ \tab & be in node $a_1$  \tab & $a_0 \rightarrow a_1$ \\ [1.2ex]
At step 2 \tab & with \tab & $P^{(2)}_{prev}$ \tab & be in node $a_0$  \tab & $a_0 \rightarrow a_1 \rightarrow a_0$ \\ [1.2ex]
          & with \tab & $P^{(2)}_{new}$ \tab & be in node $a_2$  \tab & $a_0 \rightarrow a_1 \rightarrow a_2$ \\ [1.2ex]
At step 3 \tab & with \tab & $P^{(3)}_{prev}$ \tab & be in node $a_0$  \tab & $a_0 \rightarrow a_1 \rightarrow a_0 \rightarrow a_1$ \\ [1.2ex]
if $P^{(2)}_{prev} == 1$ & with \tab & $P^{(3)}_{new}$ \tab & be in node $a_2$  \tab & $a_0 \rightarrow a_1 \rightarrow a_0 \rightarrow a_2$ \\ [1.2ex]
if $P^{(2)}_{new} == 1$ & with \tab & $P^{(3)}_{prev}$ \tab & be in node $a_1$  \tab & $a_0 \rightarrow a_1 \rightarrow a_2 \rightarrow a_1$ \\ [1.2ex]
          & with \tab & $P^{(3)}_{new}$ \tab & be in node $a_3$  \tab & $a_0 \rightarrow a_1 \rightarrow a_2 \rightarrow a_3$ \\ [1.2ex]
          & with \tab & $P^{(3)}_{already}$ \tab & be in node $a_0$  \tab & $a_0 \rightarrow a_1 \rightarrow a_2 \rightarrow a_0$ \\ [1.2ex]
\end{tabular}
\end{table}

\begin{figure}[H]
   \centering
  \includegraphics[cfbox=MyColor 0.1pt 5pt, scale=1.4]{"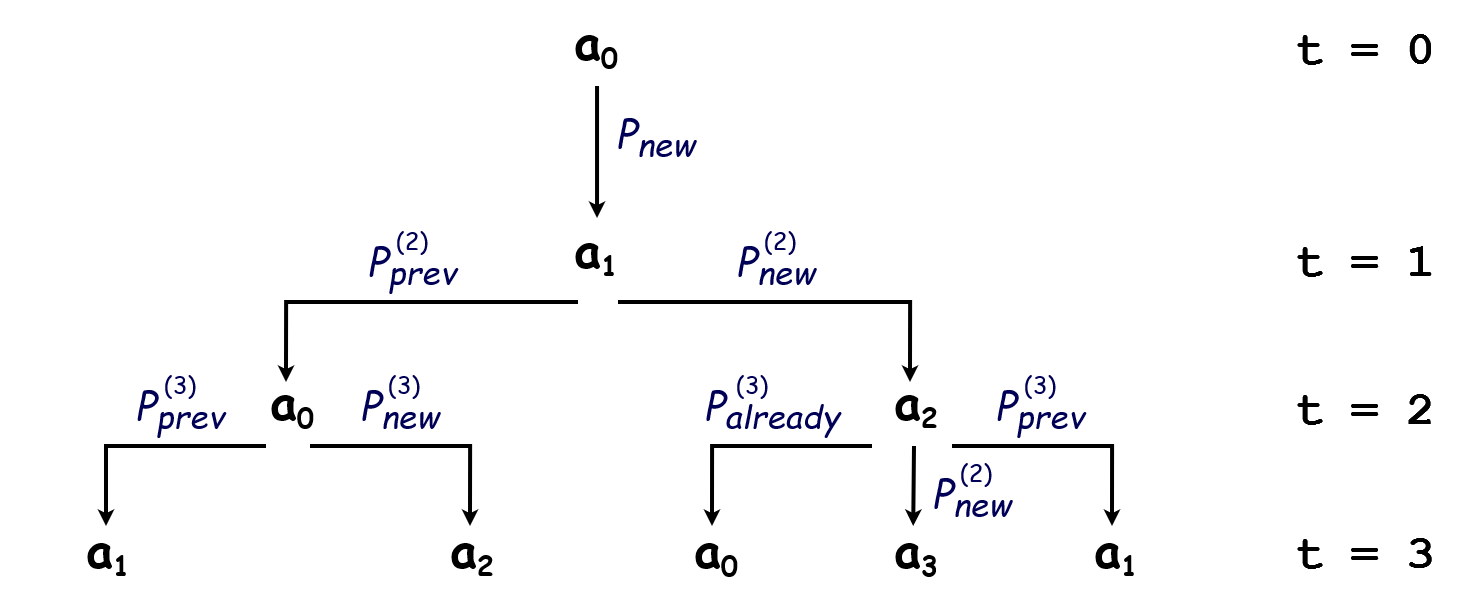"}
\caption{Shows the probabilities of visiting nodes in a three-step neighborhood.}
   \label{SampleGraph_3}
\end{figure}

\tab The tree-like representation shows the visiting probabilities of the nodes in the 3-step neighborhood (Figure~\ref{SampleGraph_3}). The probabilities of arriving nodes at each time step are summed to calculate the elements of the transition probability matrix for each node. The calculated probabilities for a three-step anonymous random walk are presented in Table~\ref{ARWProbabilities} to exemplify the idea. The Transition probability matrix, TPM, obtained for a three-step anonymous walk can be given as in Table~\ref{TPM}.

\begin{table}[H]
\centering
  \begin{tabular}{|c|c|l|c|}
    \hline
    Step &       & Probability & Node \\
    \hline
    $ 1   $ & $   P^{(1)}_{0,1}   $ & $ \frac{1}{d_0}$ & $ 0\;\rightarrow \;1 $ \\ [1.5ex]
    \hline
    $ 2   $ & $   P^{(2)}_{1,0}   $ & $  P^{(1)}_{0,1} \times \frac{1}{d_1}$ & $ 1\;\rightarrow \; 0 $ \\ [1.5ex]
            & $   P^{(2)}_{1,2}   $ & $  P^{(1)}_{0,1} \times \frac{d_1-1}{d_1}$ & $ 1\;\rightarrow \;2$ \\ [1.5ex]
    \hline
        
    $ 3 $ & $  P^{(3)}_{2,0}   $ & $  P^{(2)}_{1,2} \times \delta_{2,0}\times \frac{1}{d_2}$ & $ 2\;\rightarrow \;0$ \\ [1.5ex]
     & $  P^{(3)}_{2,1}   $ & $  P^{(2)}_{1,2} \times \frac{1}{d_2}$ & $ 2\;\rightarrow \;1$ \\ [1.5ex]
    & $  P^{(3)}_{2,3}   $ & $  P^{(2)}_{1,2} \times \left[\frac{d_2-2}{d_2} + [1 - \delta_{2,0}] \times \frac{1}{d_2}\right]$ & $ 2\;\rightarrow \;3$ \\ [1.5ex]
    &                    &                                       & \\
    & $  P^{(3)}_{0,1}   $ & $  P^{(2)}_{1,0} \times \frac{1}{d_0}$ & $ 0\;\rightarrow \;1$ \\ [1.5ex]
    & $  P^{(3)}_{0,2}   $ & $  P^{(2)}_{1,0} \times \delta_{0,2}\times \frac{1}{d_0}$ & $ 0\;\rightarrow \; 2$ \\ [1.5ex]
     & 
     $  P^{(3)}_{2,3}   $ & $  P^{(2)}_{1,0} \times \left[\frac{d_0-2}{d_0} + [1 - \delta_{0,2}] \times \frac{1}{d_0}\right]$ & $ 2\;\rightarrow \;3$ \\ [1.5ex]
    \hline
\end{tabular}
\caption{Anonymous random walk probabilities for a set of three-step  anonymous walks. Theoretical probability values describe the probabilities of the vicinity of the neighbors at each time step. Here $d_i$ is the degree of each node and $\delta_{i,j}$ is the Kronecker delta, which indicates the connection between node $v_i$, and node $v_j$. }
\label{ARWProbabilities}
\end{table}

\begin{table}[H]
\centering
\begin{tabular}{clcccclc}
                                &   & $a_0$              &   $a_1$                           &   $a_2$           & $a_3$ &         &          \\  [1.5ex] \cline{2-2}\cline{7-7}
\multicolumn{1}{c|}{}           &   & 1                  &   0                               &   0               & 0               & \multicolumn{1}{l|}{}    & $t=0$ \\ [1.5ex]

\multicolumn{1}{c|}{}    &   & 0                  &   $P^{(1)}_{0,1}$                 &   0               & 0               & \multicolumn{1}{l|}{}    & $t=1$\\ [1.5ex]
\multicolumn{1}{c|}{$TPM =$}           &   &  $P^{(2)}_{1,0}$   &   0                               &   $P^{(2)}_{1,2}$ & 0               & \multicolumn{1}{l|}{}    & $t=2$\\ [1.5ex]
\multicolumn{1}{c|}{}           &   &  $P^{(3)}_{2,0}$   &   $P^{(3)}_{0,1}+P^{(3)}_{2,1}$   &   $P^{(3)}_{0,2}$ & $P^{(3)}_{2,3}$ & \multicolumn{1}{l|}{}  & $t=3$  \\ [1.5ex]\cline{2-2} \cline{7-7} 
\end{tabular}
\caption{Transition probability matrix element for each time step.}
\label{TPM}
\end{table}

\subsection{Obtaining Transition Matrix - Implementation}
\tab The above transition probability matrix (Table~\ref{TPM}) is obtained from a given graph by creating a set of random walks starting from each network node. The elements of the probability matrix are obtained by counting the number of anonymous random walk sequences, which means how many times each node is visited at each time step.

\begin{equation}
d_{i,j} = \sum_k \delta(A^{(k)}_{i,j} , j)
\end{equation}

\tab This process provides a distribution matrix for each node. The obtained distributions are normalized to obtain transition probability matrices. The algorithm~\ref{anonymouswalk} illustrates the process of creating a set of anonymous random walks and obtaining the transition probability matrix.

\bigskip

\begin{algorithm}[H]
  \label{anonymouswalk}
\DontPrintSemicolon
  \caption{Methodology of Creating Transition Probability Matrix}
  \KwInput{\\
  \hspace{1cm} Graph $G(V;E)$ \\ 
  \hspace{1cm} Walks per node $\eta$\\
  \hspace{1cm} Walk length $m$
  }
  \KwOutput{Node representation $X(v)$ for each node $v \in V$}

  \For{each $v \in V$}    
        { 
        	Initialize $walks_v$ to Empty  \tcp*{Random walks set}
        	Initialize $anons_v$ to Empty  \tcp*{Anonymous random walks set}
        	Initialize $tpm_v = [t]_{|m+1|\times|m+1|}, t=0$ \tcp*{Transition probability matrix}
        	\For{$i=0$ to $\eta$}    
                {
                    $w=$ UniformlyRandomWalk$(G, v, m)$ \\
                    Append a walk $w$ to $walks_v$
                }
            $anons_v=$ AnonymousRandomWalk$(walks_v, v, m)$ \\
            $k=0$
            
            \tcp{Go through the steps of the anonymous random walks}
            \For{$step$ in $anons_v$}  
                {
                    \tcp{Count the number of visited node at each step of the walk}
                    Update $tpm_v[k, step]$ $+${=} 1 \\
                    Update $k$ $+${=} 1
                }
            $X(v) = normalize (tpm_v)$
        }
        Return $X$
\end{algorithm}

\bigskip

\begin{figure}[H]
   \centering
  \includegraphics[width=0.99\textwidth]{"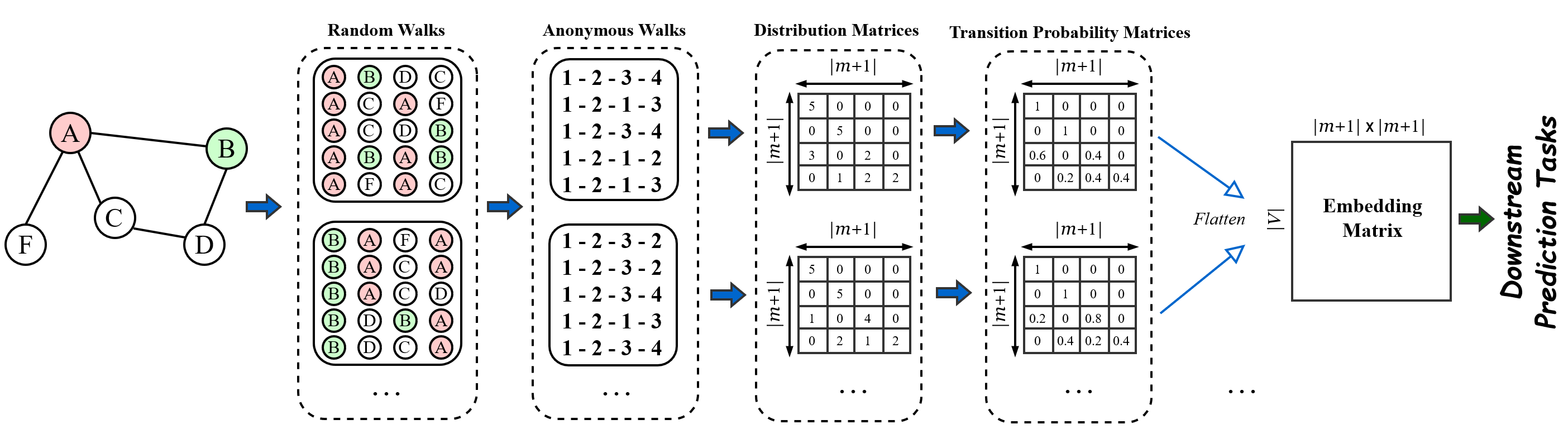"}
\caption{Architecture of the TPM model}
   \label{TPM_Framework}
  \end{figure}

\tab Once the transition probability is obtained for each node, the transition probability matrix may be used in various ways for calculating the network properties. In this work, the aim is to show how well the local characteristics of a given node are preserved in the definition of the transition probability matrix. To this end, it is used as node features in node classification, link prediction, visualization, and cross-networks generalization tasks. The implementation of the embedding process consists of four steps (Please see, Figure~\ref{TPM_Framework}):

\begin{enumerate}[(i)]
\item Create a set of random walks, $\eta$ with length $m$ starting
  from each node.
\item Convert each random walk to an anonymous walk.
\item Calculate the distribution matrix ($|m+1|\times|m+1|$) for each node.
\item Obtain the Transition Probability Matrix (TPM, $|m+1|\times|m+1|$
  matrix) for each node after normalizing the distribution matrix from the previous step.
\item Use TPM as node features in the embedding process.
\end{enumerate}

Here $m$ refers to the number of steps in a random walk. Because the purpose is to collect the local information about a node, a high number of steps causes it to go too deep into the graph.

\section{Experiments}
\tab The proposed method (TPM) is compared against numerous state-of-the-art baseline embedding techniques on four downstream tasks: node classification, link prediction, network visualization, and cross-networks generalization. In addition, parameter sensitivity is conducted. The results demonstrate that our method considerably outperforms the baseline embedding techniques in all the network analysis tasks.

\subsection{Datasets}
\tab Three different citation networks are employed to test the quality of the proposed embedding method (TPM). Table~\ref{Networks} shows the network statistics of Cora~\cite{mccallum2000automating}, Citeseer~\cite{giles1998citeseer}, and Pubmed~\cite{sen2008collective} networks, which are commonly used for testing algorithms and predicting network properties in a considerable number of publications~\cite{molokwu2020node, xie2020multi, mohammed2021degree, buffelli2022impact}. Cora, Citeseer, and Pubmed comprise academic publications, which are the nodes, and citation relations between papers, which are the edges. Each node is labeled according to its research field. All of the datasets are open to the public \footnote{\url{https://linqs.soe.ucsc.edu/data} [Accessed date: 12 March 2022]}.

\begin{table}[h]
\centering
\caption{Overview of three citation networks datasets.}
\label{Networks}
\begin{tabular*}{\columnwidth}{
  @{\extracolsep{\fill}\hspace{\tabcolsep}}
  lccc
  @{\hspace{\tabcolsep}}
}
\hline
            & Cora      & CiteSeer      & Pubmed    \\ \hline
Nodes       & 2,708     & 3,327         & 19,717    \\ 
Edges       & 5,429     & 4,732         & 44,338    \\ 
Classes     & 7         & 6             & 3         \\ \hline

\end{tabular*}%
\end{table}

\subsection{Baselines}
\tab We compare the performance of our proposed TPM model with six well-established feature extraction algorithms (DeepWalk~\cite{perozzi2014deepwalk}, node2vec~\cite{grover2016node2vec}, LINE~\cite{tang2015line}, GCN-Graph Convolutional Network~\cite{welling2016semi}, GraphSAGE~\cite{hamilton2017inductive}, and GAT~\cite{velickovic2017graph}).

\begin{itemize}
  \item \textbf{DeepWalk}~\cite{perozzi2014deepwalk} is an unsupervised approach that learns low-dimensional representations of nodes by using local structural information collected from truncated random walks.
  \item \textbf{node2vec}~\cite{grover2016node2vec} proposes a more flexible strategy (biased random walks) for sampling node sequences, which allows it to better balance the importance of the local and global structure of a graph. 
  \item \textbf{LINE}~\cite{tang2015line} optimizes a predefined objective function in a way that maintains the network topology on both the local (\textit{first-order} proximity) and the global (\textit{second-order} proximity) levels. For each node, LINE computes two feature representations and then utilizes an efficient method to combine the vector representations learned by LINE (\textit{first-order}) and LINE (\textit{second-order}) into a single, $d$-dimensional vector. 
  \item \textbf{GCN-Graph Convolutional Network}~\cite{welling2016semi} is a widely common form of Graph Neural Networks (GNNs). In GCN, the representation of each node is created by a convolution layer that aggregates both the node's attributes and the attributes of its surrounding nodes. The success of the GCN is attributable to two techniques: first, the adjacency matrix is normalized by the degrees, and second, each node is given a self-connection. 
  \item \textbf{GraphSAGE}~\cite{hamilton2017inductive} is the pioneering inductive framework, and it uses the concept of spatial GCN-Graph Convolutional Network to learn node representations. Essentially, it creates an aggregation function (might be mean (as in GCN), pooling, or LSTM) to collect attributes from each node's neighbors.
  \item \textbf{GAT}~\cite{velickovic2017graph} combines techniques of attention with Graph Neural Networks (GNN) in an effort to improve its learning capability in relation to the characteristics of neighborhoods. 
\end{itemize}
 
\tab According to the relevant literature, these baselines have exhibited state-of-the-art embedding performances on the datasets used in our tests. As for parameter settings, the number of walks per node, length of walks, window size, and negative sampling are set to 10, 80, 30, and 10 for both DeepWalk and node2vec, respectively. We observed experimentally that setting $p=2$ and $q=1$ in node2vec produces superior results across all datasets. We used the recommended hyperparameters and default architectures for the rest baseline methods based on the corresponding original papers. We set the number of walks per node $\eta=20$ and the length of walks $m=10$ for our proposed (TPM) model. For the sake of fair comparison, the representational dimension ($d$) for all baselines has been fixed to 128.

\subsection{Evaluation Metrics}
\tab As evaluation metrics, we use Macro-F1 and Micro-F1 scores~\cite{zaki2014data} for the node classification task, and we use AUC (area under the ROC curve)~\cite{hanley1982meaning} for the link prediction task. Higher values of these metrics indicate better performance of the related network embedding technique. 

\begin{itemize}
  \item \textbf{Macro-F1} is the mean of the per-class F1 scores.
  \item \textbf{Micro-F1} calculates a global average F1 score by adding the total number of True Positives (TP), False Negatives (FN), and False Positives (FP) across all labels.
\end{itemize}

The formulas of Macro-F1 and Micro-F1 are:

\begin{equation}
\text { Macro-F1} =\frac{1}{c} \sum_{i=1}^c \frac{2 T P_i}{2 T P_i+F P_i+F N_i}
\end{equation}

\begin{equation}
\text { Micro-F1} =\frac{2 \sum_{i=1}^c T P_i}{\sum_{i=1}^c 2 T P_i+F P_i+F N_i} 
\end{equation}
where $c$ denotes the number of classes.

\begin{itemize}
  \item Area under the ROC curve, or AUC, is the probability that the score of a positive link (existent) is greater than the score of a randomly selected link that does not exist. AUC compares $n$ times the scores of a randomly picked positive (existent) and negative (nonexistent) link. If $n^{\prime}$ times an existent link's score is greater than a nonexistent link's, and $n^{\prime \prime}$ times they share the same score, the AUC is:
\end{itemize}
\begin{equation}
\text { AUC} =\frac{n^{\prime}+0.5 n^{\prime \prime}}{n}
\end{equation}

\begin{table}[h]
\caption{Summary of node classification results. Bold values mean the
  best results.}
\begin{tabular*}
{\columnwidth}{
  @{\extracolsep{\fill}\hspace{\tabcolsep}}
  l cc cc cc
  @{\hspace{\tabcolsep}}
}
\hline
\multirow{2}{*}{} & \multicolumn{2}{c}{CORA}       & \multicolumn{2}{c}{CiteSeer}   & \multicolumn{2}{c}{Pubmed}     \\ \cline{2-7} 
 &
  \multicolumn{1}{l}{Micro-F1} &
  \multicolumn{1}{l}{Macro-F1} &
  \multicolumn{1}{l}{Micro-F1} &
  \multicolumn{1}{l}{Macro-F1} &
  \multicolumn{1}{l}{Micro-F1} &
  \multicolumn{1}{l}{Macro-F1} \\ \hline
DeepWalk          & 0.715          & 0.692          & 0.560           & 0.521          & 0.659          & 0.652          \\
node2vec          & 0.846          & 0.838          & 0.751          & 0.703          & 0.722          & 0.717          \\
LINE              & 0.797          & 0.793          & 0.582          & 0.536          & 0.728          & 0.723          \\
GCN               & 0.819          & 0.807          & 0.714          & 0.680           & 0.788          & 0.781          \\
GSAGE             & 0.829          & 0.813          & 0.684          & 0.661          & 0.760           & 0.754          \\
GAT               & 0.834          & 0.830           & 0.729          & 0.688          & 0.795          & 0.789          \\ \hline
TPM               & \textbf{0.858} & \textbf{0.849} & \textbf{0.781} & \textbf{0.743} & \textbf{0.821} & \textbf{0.818} \\ \hline
\end{tabular*}%
\label{NodeClassificationACC}
\end{table}

\subsection{Node Classification}
\tab Node classification is the process of classifying unlabeled nodes based on their proximity to nodes whose classes are already known (called labeled nodes). The node features embedding data are divided into test and training data sets. Two train and test sets consist of $80\%$ training data and $20\%$ for test data for each of the three networks. To solve the problem of imbalanced target classes, 10-fold cross-validation is performed, and the experiment results are a 10-run average. As the classification method, Logistic Regression, Support Vector Machine (SVM), Decision Tree, and Multi-Layer Perceptron (MLP) algorithms have been used for testing and comparison purposes using the default settings for the scikit-learn Python package. The results of the above-mentioned classification algorithms are within the error limits. Hence, only the results obtained by using Multi-Layer Perceptron are presented in this section. Table~\ref{NodeClassificationACC} shows the classification results of all seven embedding algorithms obtained by using a three-layer neuron network. The Multi-Layer Perceptron model consists of three dense layers of $128$, $64$ neurons with ReLU activation, while in the final layer number of neurons is chosen according to the number of classes of the given network with softmax activation. Adam optimizer and the categorical\_crossentropy loss function employed in the Multi-Layer Perceptron. Micro-F1 and Macro-F1 were used as the accuracy measure. Table~\ref{NodeClassificationACC} exhibits comparative results obtained using six well-tested node feature extraction methods and the proposed transition probability matrix, TPM, based feature extraction algorithm. TPM-based feature vectors performed better in classifying the nodes for all three networks in terms of both Micro-F1 and Macro-F1.

\subsection{Link Prediction}
\tab Link prediction is the task of predicting the probability of being connected between the nodes. For example, in social networks, which is connected with whom; in citation networks, who is co-author with whom; in biological networks, which genes or proteins interact with, are some crucial areas in the literature. Once the proposed embedding methodology has proved successful for node classification, it has also been used for link prediction. To create the link representation $g(u, v)$, we use binary vector operators between two given nodes $u$ and $v$. In the present work, the node embedding vectors are used together with four amalgamation operators introduced in the literature~\cite{grover2016node2vec}. These four combination techniques keep the link vector's dimension the same as the node's, improving computing efficiency. The operators, Average, Hadamard, Weighted-L1, and Weighted-L2 have been tested for the best result. The use of the Hadamard operator has given the best predictions. The formula of Hadamard operation is: 

\begin{equation}
g_i(u, v)=[f(u) \cdot f(v)]_i=f_i(u) * f_i(v)
\end{equation}

\bigskip

where $f(u)$ and $f(v)$ are the vectors of the pair $(u, v)$, $f_i(u)$ denotes the $i$th component of $f(u)$, and $g_i(u, v)$ represents the $i$th component of the link representation $g(u, v)$. 

\bigskip

\begin{table}[h]
\centering
\caption{Summary of link prediction results measured by AUC. Bold
  values mean the best results.}
\label{LinkPrediction}
\begin{tabular*}{\columnwidth}{
  @{\extracolsep{\fill}\hspace{\tabcolsep}}
 lcccc
  @{\hspace{\tabcolsep}}
}

\hline
         & CORA           & CiteSeer       & Pubmed         \\ \hline
DeepWalk & 0.851          & 0.824          & 0.864          \\
node2vec & 0.927          & 0.941          & 0.912          \\
LINE     & 0.856          & 0.803          & 0.833          \\
GCN      & 0.918          & 0.889          & 0.957          \\
GSAGE    & 0.910           & 0.877          & 0.931          \\
GAT      & 0.908          & 0.914          & 0.924          \\ \hline
TPM      & \textbf{0.959} & \textbf{0.962} & \textbf{0.969} \\ \hline
\end{tabular*}%
\end{table}

\tab In this part, we perform binary classification on three citation networks, i.e., Cora, Citeseer, and Pubmed. As the prediction method, Multi-Layer Perceptron (MLP) algorithm has been used for testing and comparison purposes. The Multi-Layer Perceptron model consists of three dense layers of $128$, $64$ neurons with ReLU activation, while in the final layer sigmoid activation function is employed. All experiments are performed 10 times, and the average AUC is reported. For link prediction at each batch,

\begin{enumerate}[(i)]
\item $10\%$ of the positive links (existing), randomly selected
  as the test set, the same number of negative links (non-existing)
  chosen from the original graph.
\item The remaining positive and negative links are used as the
  training set.
\end{enumerate}

Table~\ref{LinkPrediction} gives the comparisons of the link
prediction results. Transition probability matrix-based feature
vectors, TPM exhibits comparatively good results.

\subsection{Visualization}
\begin{figure}[h]
  \centering
\begin{subfigure}{.32\textwidth}
  \centering
  \includegraphics[width=1\linewidth]{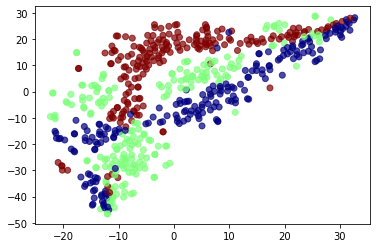}  
  \caption{DeepWalk}
  \label{Vis_DW}
\end{subfigure}
\begin{subfigure}{.32\textwidth}
  \centering
  \includegraphics[width=1\linewidth]{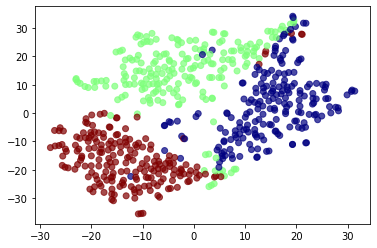}  
  \caption{GCN}
  \label{Vis_GCN}
\end{subfigure}
\begin{subfigure}{.32\textwidth}
  \centering
  \includegraphics[width=1\linewidth]{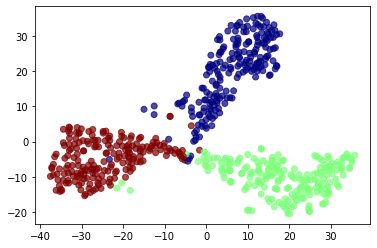}  
  \caption{TPM}
  \label{Vis_TPM}
\end{subfigure}
\caption{Visualization of 2D representations for LFR network.}
\label{Visualization}
\end{figure}

\tab Visualization is one of the best methods for testing the success of embedding algorithms. Using a controlled experiment approach, objective discrimination between the algorithms can be realized. To start the experiment, a common approach is to use well-known networks, such as Zachary Karate Club~\cite{girvan2002community} or Cora~\cite{mccallum2000automating}, for comparison purposes. Despite the common usage of these networks, real-world networks exhibit very different characteristic heterogeneous distributions of community sizes and node degrees. For this reason, instead of using any one of the commonly used networks, a network creation algorithm Lancichinetti–Fortunato-Radicchi~\cite{lancichinetti2008benchmark} (LFR algorithm), which is particularly designed for testing embedding algorithms is used. The TPM algorithm was applied to a network of 3 communities, 600 nodes, and 1334 edges. LFR algorithm has an extra parameter $\mu$ that controls the "noise". The "Noise" parameter, which takes values between 0 and 1, is an indicator of the heterogeneity of the network. In this work, the heterogeneity parameter is taken as $\mu=0.2$. The visualization results of two well-established embedding algorithms, DeepWalk, and GCN, are compared for visual inspection with  the proposed algorithm (TPM). For dimensional reduction, t-Distributed Stochastic Neighbor Embedding (t-SNE) visualization tool is used. Figure~\ref{Visualization} shows that our model is capable of producing more compact and distinct clusters than the other two methods. From Figure~\ref{Visualization}, we observe that the representations generated by the DeepWalk method have multiple clusters that overlap with each other. The GCN model provides a little more coherent representation because it transfers the attributes of the neighboring nodes through the connectivity structure to capture additional global structural data. The visual inspection of Figure~\ref{Vis_TPM} indicates that the proposed embedding algorithm (TPM) has distinct separation among three classes of the created network.

\subsection{Cross-networks generalization: Scale-free networks}
\tab Scale-free networks have a unique position since they constitute most real-world networks~\cite{barabasi2003scale}. They exhibit power-law degree distributions due to their heavy-tailed character. As shown in the degree-distribution graphic there are a limited number of nodes called hubs with a high degree and many numbers of nodes with a low degree. Most of the real-world networks show similar behavior and have scale-free degree distribution. For example, in social networks, there are a few users with many followers and many users with an average number of followers. Similarly, in airport networks, there are a few numbers of hub airports that have flights (have connections) to many different airports and many small airports which have flights to limited other airports. Therefore it is important to work and test the proposed algorithm, TPM, for scale-free networks.

\tab Therefore, we perform the task of cross-networks generalization,  which is learning information from networks of similar connectivity structures to make predictions on new networks. In this experiment, we conduct a link prediction task on networks previously unseen  during training. We generate 12 Barabási-Albert networks (BA) with the following parameters: $n=\{1000, 10000\}$ (number of nodes) and $\alpha=6$ (how many existing nodes may be attached to a new one). We train each embedding method on ten networks and then calculate the average AUC score on two test networks. We compare the performance of our proposed TPM model with four feature extraction algorithms (DeepWalk, node2vec, GCN, and GraphSAGE), and the experimental results are shown in Table~\ref{cross_networks_generalization}. For all random walk-based models (TPM, DeepWalk, and node2vec), we run a total of $50$ random walks of $10$ steps for each node (5 random walks from each generated training Barabási-Albert network). For all graph convolutional networks-based  models (GCN and GraphSAGE), we perform information aggregation per node evenly from ten generated training Barabási-Albert networks. Moreover, in the generated Barabási-Albert networks, nodes lack features. For features, we use node degrees and an embedding weight that is updated for each node  during training. Table~\ref{cross_networks_generalization} shows that  TPM significantly outperforms all four feature extraction algorithms. These results demonstrate that our TPM model well captures the local structural patterns of nodes, even in different networks having similar topologies.

\begin{table}[h]
\centering
\caption{Summary of results in terms of cross-networks generalization for two scale-free networks. Bold values mean the best results.}
\label{cross_networks_generalization}
\begin{tabular*}{\columnwidth}{
  @{\extracolsep{\fill}\hspace{\tabcolsep}}
  lccccc
  @{\hspace{\tabcolsep}}
}

\hline
                                    & DeepWalk      & node2vec       & GCN         & GSAGE      & TPM   \\ \hline
Scale-free networks ($n=1,000$)   & 0.611         & 0.664          & 0.702       & 0.713      & \textbf{0.746} \\ 
Scale-free networks ($n=10,000$)  & 0.620         & 0.659          & 0.719       & 0.717      & \textbf{0.751} \\ \hline

\end{tabular*}%
\end{table}

~\subsection{Parameter Sensitivity}
\begin{figure}[ht]
  \centering
\begin{subfigure}{.45\textwidth}
  \centering
  \includegraphics[width=1\textwidth]{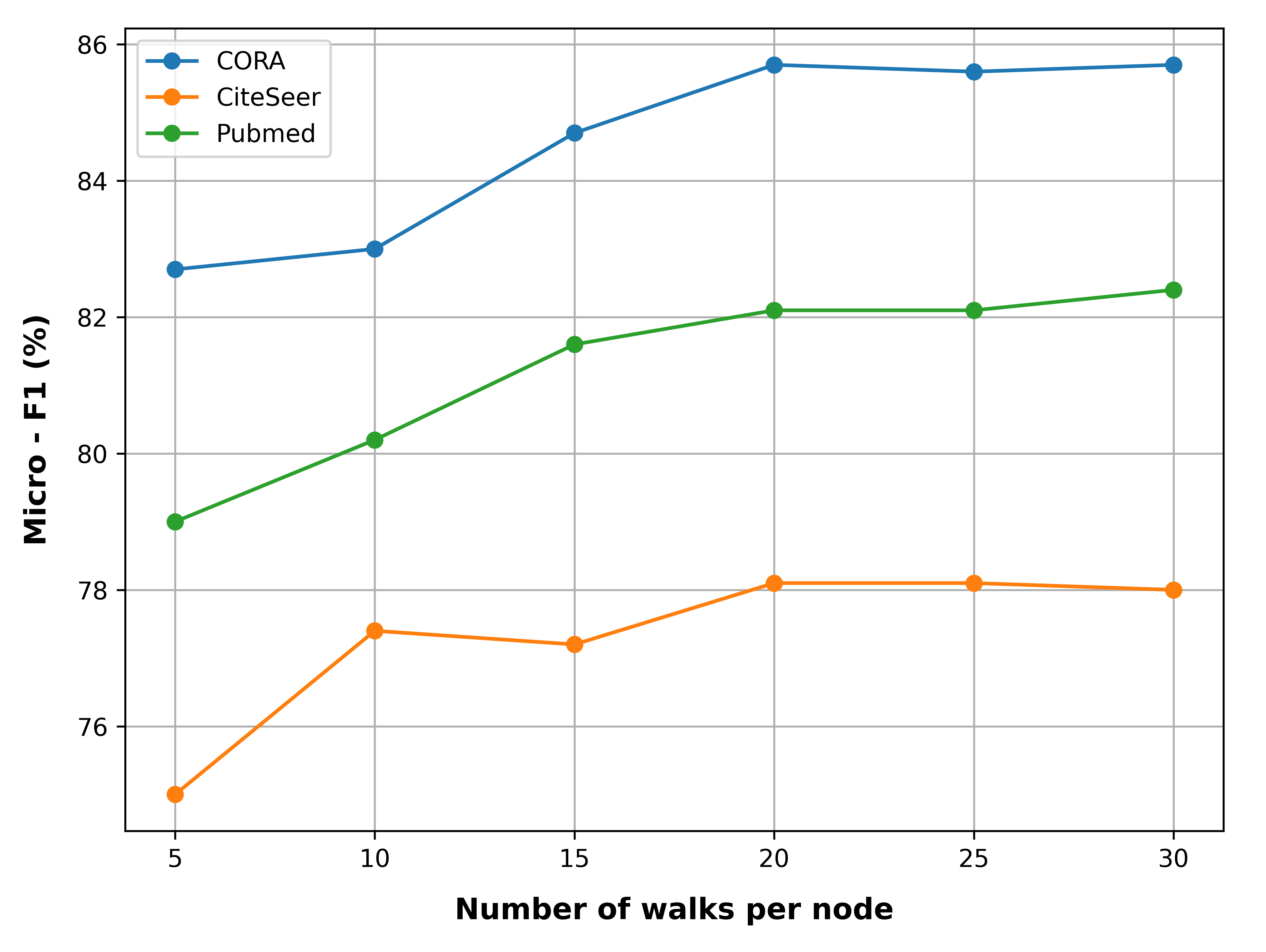}  
  \caption{}
  \label{PS_NW}
\end{subfigure}
\begin{subfigure}{.45\textwidth}
  \centering
  \includegraphics[width=1\textwidth]{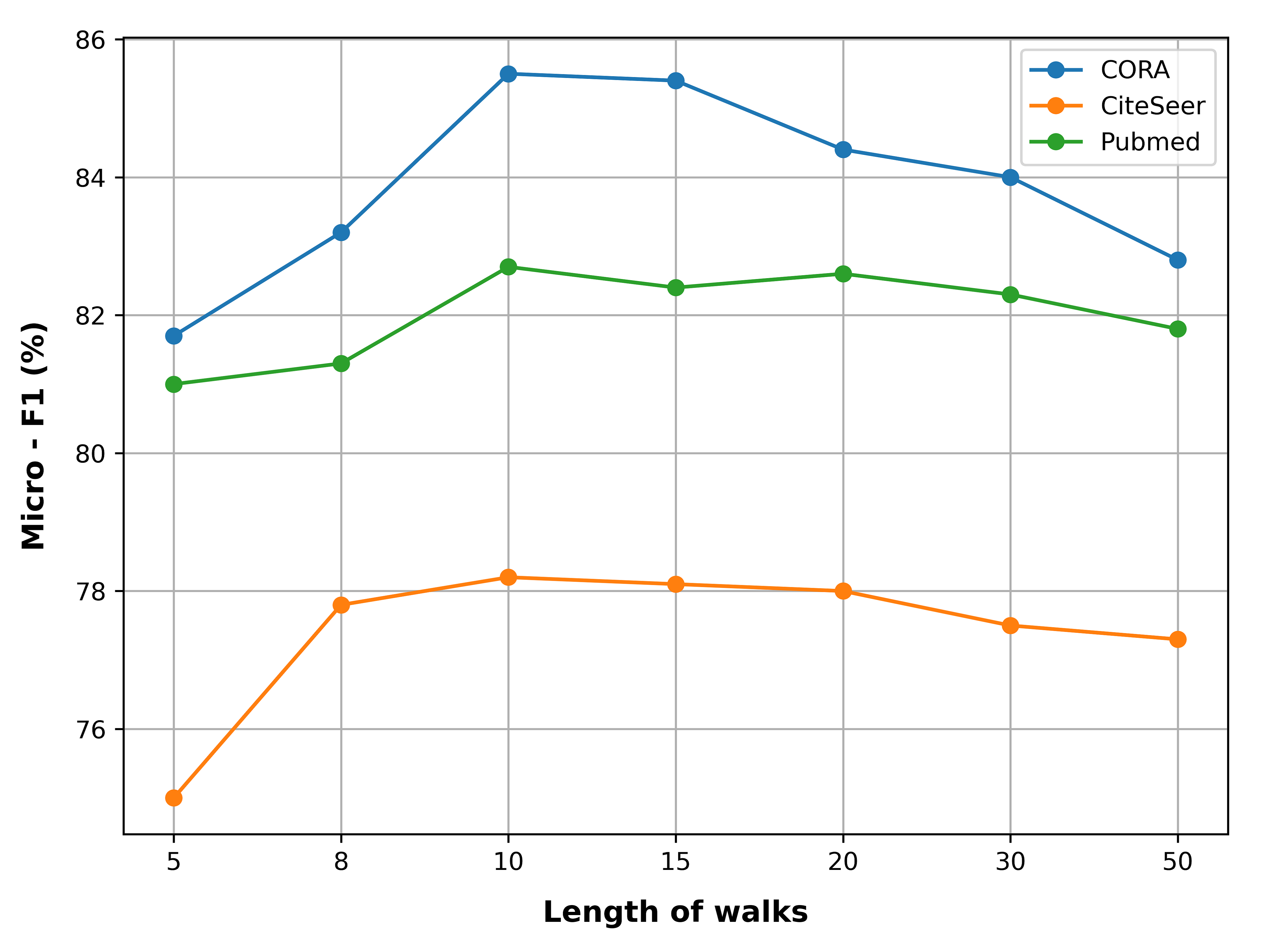}  
  \caption{}
  \label{PS_WL}
\end{subfigure}

\caption{Results of parameter experiment. ($a$) Effect of the number of walks $\eta$ on three citation networks. ($b$) Effect of the length of walks $m$ on three citation networks. }
\label{Parameter_Sensitivity}
\end{figure}

\tab The proposed embedding algorithm, TPM, includes two essential hyper-parameters, $\eta$, and $m$, the number of walks per node and the length of walks. The Micro-F1 score of node classification over Cora, Citeseer, and Pubmed with various $\eta$ and $m$ demonstrate the impact of the hyper-parameters in TPM. The number of walks per node $\eta$ and length of walks $m$ were set to $\{5, 10, 15, 20, 25, 30\}$ and $\{5, 8, 10, 15, 20, 30, 50\}$. The training ratio was set to $80\%$, and the results can be seen in Figure~\ref{Parameter_Sensitivity}. As shown in Figure~\ref{PS_NW}, the performance of the proposed model is relatively steady over the three citation networks when $\eta$ varies from $20$ to $30$. Hence, the larger the number of walks, the better the model's performance since the characteristic connectivity structure of the region around the given node is captured more thoroughly.  In Figure~\ref{PS_WL}, the proposed embedding algorithm TPM achieves the best Micro-F1 score over the three citation networks when $m$ equals $10$. The score gradually decreases with increasing walk length because the walker may move away from the node's neighborhood when the walk length $m$ is too large, thereby failing to incorporate local structural patterns properly into the walk statistics. Consequently, the parameters $\eta$ and $m$ must be tuned appropriately for various applications.

\section{Conclusions}
\tab The traditional machine learning frameworks can only be applied to networks, by mapping high-dimensional information contained in the network, to low-dimensional vector spaces. Hence, a vector representation of each node, called node embedding, is an essential initial step for processing the data obtained from networks. The embedding algorithm must represent the graph's connectivity structure, which requires mixing properties of nodes and edges. Most of the existing representation learning techniques concentrate only on the local structure of nodes, hence lacking representation of local structural patterns in downstream network analysis tasks.

\tab The present study introduces a new, scalable unsupervised node embedding algorithm that inherently contains the local connectivity structure of the nodes. Moreover, node similarities are also part of the embedding representation since the node embeddings overlap. The proposed algorithm uses anonymous walks for structural node embeddings. Starting from a given node, each walk collects local structural information in a predetermined neighborhood radius. A unique transition probability matrix represents each node in the network. Elements of each transition probability matrix consist of the probability of reaching the neighboring nodes starting from the original. Hence the transition probability matrices of neighboring nodes overlap. The overlapping transition probability matrices ensure the correct similarity measures between the neighboring nodes. The significant advantage of the proposed method is to capture local structural patterns rather than the identity of a node by visiting a limited number of close neighbors using short random walks. This reduces computation time, improves performance, and gathers information without going deep into a graph. Moreover, the transition probability matrix method has superior prediction potential in identifying similar connectivity structures in remote network parts. This possibility extends the use of the proposed embedding vectors, created using the information of a given network, onto unstudied networks. The flattened transition probability matrix elements are used as node feature vectors in the present work. Experiments on three commonly used real-world networks and synthetic networks, presented in the experiments section, demonstrated the effectiveness of the proposed method (TPM). For future work, the extensibility of TPM on attributed networks and temporal dynamic networks will be considered.

\bibliographystyle{unsrt}
\bibliography{references}

\end{document}